\documentclass[12pt]{article}

\usepackage{amsmath,amssymb,amsfonts}
\usepackage{array}
\usepackage{algorithm2e}
\usepackage{tabularx}
\usepackage{subcaption}
\usepackage{graphicx}
\usepackage[biblabel]{cite}
\usepackage[colorlinks=true, linkcolor=blue, citecolor=blue, urlcolor=blue]{hyperref}

\usepackage[a4paper,margin=25mm]{geometry}

\begin{document}

\title{QGEC : Quantum Golay Code Error Correction}

\author{
    Hideo Mukai\thanks{Computer Science Program, Graduate School of Science and Technology, Meiji University, 1-1-1 Higashimita, Tama-ku, Kawasaki, Kanagawa 214-8571, Japan; Department of Computer Science, School of Science and Technology, Meiji University, 1-1-1 Higashimita, Tama-ku, Kawasaki, Kanagawa 214-8571, Japan. Corresponding Author, e-mail: mukai@meiji.ac.jp}
    \and
    Hoshitaro Ohnishi\thanks{Computer Science Program, Graduate School of Science and Technology, Meiji University, 1-1-1 Higashimita, Tama-ku, Kawasaki, Kanagawa 214-8571, Japan} 
}

\date{}

\maketitle

\begin{abstract}
Quantum computers have the possibility of a much reduced calculation load compared with classical computers in specific problems. Quantum error correction (QEC) is vital for handling qubits, which are vulnerable to external noise. In QEC, actual errors are predicted from the results of syndrome measurements by stabilizer generators, in place of making direct measurements of the data qubits. 

Here, we propose Quantum Golay code Error Correction (QGEC), a QEC method using Golay code, which is an efficient coding method in classical information theory. We investigated our method's ability in decoding calculations with the Transformer. We evaluated the accuracy of the decoder in a code space defined by the generative polynomials with three different weights sets and three noise models with different correlations of bit-flip error and phase-flip error. Furthermore, under a noise model following a discrete uniform distribution, we compared the decoding performance of Transformer decoders with identical architectures trained respectively on Golay and toric codes. 

The results showed that the noise model with the smaller correlation gave better accuracy, while the weights of the generative polynomials had little effect on the accuracy of the decoder. In addition, they showed that Golay code requiring 23 data qubits and having a code distance of 7 achieved higher decoding accuracy than toric code which requiring 50 data qubits and having a code distance of 5. This suggests that implementing quantum error correction using a Transformer may enable the Golay code to realize fault-tolerant quantum computation more efficiently. 

\end{abstract}

\section{Introduction}
Quantum computation is performed using qubits as the smallest units of information. It is a computational paradigm that has the potential to provide solutions rapidly for certain problems that would require an enormous amount of time to solve using classical computation. However, qubits are inherently vulnerable to external noise due to the problem of decoherence \cite{zurek2003}.

To perform accurate quantum computation, it is essential to employ quantum error correction techniques that can detect and properly correct errors occurring on qubits. Similar to error correction in classical computation, redundancy is introduced to the information to be preserved through encoding. However, in quantum computers, one must consider the problem more carefully, since the no-cloning theorem \cite{wootters1982} proves that there exists no unitary operation capable of creating an identical copy of the state of a given qubit.

Golay code \cite{golay1949} is an efficient code in classical error correction with high error tolerance and relatively low redundancy. It is a perfect code, meaning that when considering Hamming spheres of radius three centered on each word, the spheres cover all possible words without overlapping \cite{macwilliams1977}. It can also be applied to quantum error correction \cite{quantumgolay_database}. In this case, by defining two parity-check matrices, the generators of the X- and Z-type stabilizers correspond to them, and the code space is constructed as a Calderbank-Shor-Steane (CSS)-type stabilizer code \cite{calderbank1998}.

Many approaches based on machine learning have been proposed for decoding quantum error-correcting codes. In particular, for decoding surface codes such as toric code \cite{kitaev2003}, various decoder architectures have been developed to effectively capture their geometric structures \cite{Wu2024}\cite{FjelddahlBengtsson2024}\cite{fitzek2020}\cite{varsamopoulos2017}\cite{varbanov2025}\cite{hu2025}. Beyond simple multilayer perceptrons, studies have reported that architectures such as convolutional neural networks (CNNs) and graph neural networks (GNNs) are well suited to specific cases, and that hybrid architectures incorporating Transformers have achieved favorable logical error rates \cite{Krastanov2017}\cite{Wang2022}\cite{Lange2023}\cite{Wang2023}. 

In this study, we performed decoding of [[23,1,7]] Golay code by using a Transformer-based approach and investigated how the decoding accuracy is affected by the choice of generator polynomials and noise models. Three generator polynomials with weights of 8, 12, and 16 were selected, and by fixing the noise model as a discrete uniform distribution, we examined how the density of the parity-check matrix affects the decoding performance. Furthermore, by fixing the generator polynomial with weight 8 and keeping the parity-check matrix unchanged while varying the degree of correlation between bit-flip and phase-flip errors, we analyzed how the correlation structure of the noise model impacts the decoding accuracy. Finally, under a noise model following the same discrete uniform distribution, we compared the decoding performance of Transformer decoders with identical architectures trained respectively on Golay code defined by a generator polynomial of weight 8 and on toric code. 

The results showed that, across all three noise models with different correlations between bit-flip and phase-flip errors, there was no significant difference in decoding accuracy depending on the choice of generator polynomial. In addition, for all three generator polynomials with different weights, it was observed that lower correlations between X and Z errors led to higher decoding accuracy. In the comparison between Golay and toric codes under a noise model following a discrete uniform distribution, it was found that when the physical error rate was 5\%, the logical error rate of Golay code was approximately 40\% lower than that of toric code.

\section{Method} 
\label{2}
We defined parity check matrices of [[23,1,7]] Golay code with three generative polynomials having different weights. Then, we evaluated the accuracy of the decoders trained by Transformer, while changing the correlation between bit-flip error and phase-flip error. We adapted the encoder block of Transformer, which learned the problem of decoding [[23,1,7]] Golay code as a regression task.
\subsection{Generator polynomial and parity check matrix}
\label{2.1}
In Golay code, there is arbitrariness in the choice of generative polynomials and corresponding parity check matrices. Here, we chose three generative polynomials of weight 8, 12, and 16 for [[23,1,7]] Golay codes. 
\begin{equation}
    h_1(x) = x^{12} + x^{10} + x^{7} + x^4 + x^3 + x^2 + x + 1
\end{equation}
\begin{equation}
    h_2(x) = x^{16} + x^{14} + x^{12} + x^{11} + x^{10} + x^8 + x^6 + x^5 + x^3 + x^2 + x + 1
\end{equation}
\begin{equation}
    h_3(x) = x^{21} + x^{18} + x^{17} + x^{16} + x^{15} + x^{14} + x^{13} + x^{12} + x^{11} + x^{10} + x^8 + x^7 + x^5 + x^3 + x + 1
\end{equation}

The coefficients of each degree in these polynomials were formed into a 23-bit sequence. The circularly shifted bit sequences were regarded as elements in a vector space of 23 dimensions in $F_2$. Then, the parity check matrix was obtained from the 11 independent elements among the 23. We composed the following parity check matrices from the polynomials.
\begin{equation}
    H_1 = 
\left[
\begin{array}{ccccccccccccccccccccccc}
  1 &1 &1 &1 &1 &0 &0 &1 &0 &0 &1 &0 &1 &0 &0 &0 &0 &0 &0 &0 &0 &0 &0  \\
  0 &1 &1 &1 &1 &1 &0 &0 &1 &0 &0 &1 &0 &1 &0 &0 &0 &0 &0 &0 &0 &0 &0  \\
  0 &0 &1 &1 &1 &1 &1 &0 &0 &1 &0 &0 &1 &0 &1 &0 &0 &0 &0 &0 &0 &0 &0  \\
  0 &0 &0 &1 &1 &1 &1 &1 &0 &0 &1 &0 &0 &1 &0 &1 &0 &0 &0 &0 &0 &0 &0  \\
  0 &0 &0 &0 &1 &1 &1 &1 &1 &0 &0 &1 &0 &0 &1 &0 &1 &0 &0 &0 &0 &0 &0  \\
  0 &0 &0 &0 &0 &1 &1 &1 &1 &1 &0 &0 &1 &0 &0 &1 &0 &1 &0 &0 &0 &0 &0  \\
  0 &0 &0 &0 &0 &0 &1 &1 &1 &1 &1 &0 &0 &1 &0 &0 &1 &0 &1 &0 &0 &0 &0  \\
  0 &0 &0 &0 &0 &0 &0 &1 &1 &1 &1 &1 &0 &0 &1 &0 &0 &1 &0 &1 &0 &0 &0  \\
  0 &0 &0 &0 &0 &0 &0 &0 &1 &1 &1 &1 &1 &0 &0 &1 &0 &0 &1 &0 &1 &0 &0  \\
  0 &0 &0 &0 &0 &0 &0 &0 &0 &1 &1 &1 &1 &1 &0 &0 &1 &0 &0 &1 &0 &1 &0  \\
  0 &0 &0 &0 &0 &0 &0 &0 &0 &0 &1 &1 &1 &1 &1 &0 &0 &1 &0 &0 &1 &0 &1  \\
\end{array}
\right]
\end{equation}
\begin{equation}
    H_2 = 
\left[
\begin{array}{ccccccccccccccccccccccc}
  1&0&1&1&1&0&1&1&1&1&1&1&0&1&1&0&0&0&0&0&0&0&0  \\
  0&1&0&1&1&1&0&1&1&1&1&1&1&0&1&1&0&0&0&0&0&0&0  \\
  1&1&1&1&0&1&1&0&1&0&1&1&1&0&1&0&1&0&0&0&0&0&0  \\
  1&1&1&0&1&0&0&1&1&0&0&1&1&1&1&1&1&0&0&0&0&0&0  \\
  0&0&1&0&1&1&1&0&1&1&1&1&1&1&0&1&1&0&0&0&0&0&0  \\
  1&1&1&1&1&1&1&0&1&1&1&0&0&0&0&1&0&1&0&0&0&0&0  \\
  1&1&1&1&1&0&1&0&1&1&0&0&1&1&0&0&1&0&1&0&0&0&0  \\
  1&1&1&1&1&0&0&0&1&1&0&1&1&0&1&0&0&1&0&1&0&0&0  \\
  1&1&1&1&1&0&0&1&1&1&0&1&0&0&0&1&0&0&1&0&1&0&0  \\
  1&1&1&1&1&0&0&1&0&1&0&1&0&1&0&0&1&0&0&1&0&1&0  \\
  1&1&1&1&1&0&0&1&0&0&0&1&0&1&1&0&0&1&0&0&1&0&1  \\
\end{array}
\right]
\end{equation}
\begin{equation}
    H_3 = 
\left[
\begin{array}{ccccccccccccccccccccccc}
  1&0&1&1&1&1&1&1&1&1&0&1&1&0&1&1&1&1&1&0&0&0&0  \\
  1&1&1&0&0&1&1&1&1&1&1&1&1&1&1&1&0&1&0&1&0&0&0  \\
  1&1&1&1&1&0&1&1&0&0&1&1&1&1&1&0&1&1&1&1&0&0&0  \\
  0&1&0&1&1&1&1&1&1&1&1&0&1&1&0&1&1&1&1&1&0&0&0  \\
  0&1&1&1&0&0&1&1&1&1&1&1&1&1&1&1&1&0&1&0&1&0&0  \\
  1&1&0&1&1&1&1&1&0&1&1&1&0&1&1&1&0&1&1&0&1&0&0  \\
  1&1&1&0&1&1&1&0&1&0&1&0&1&1&1&1&1&1&1&0&1&0&0  \\
  0&1&1&1&1&1&0&1&1&0&0&1&1&1&1&1&0&1&1&1&1&0&0  \\
  1&0&1&1&1&0&1&0&1&1&1&1&1&1&0&1&0&1&1&1&1&0&0  \\
  1&1&1&1&1&1&1&0&1&0&0&1&1&1&0&1&1&1&0&1&0&1&0  \\
  1&1&0&1&0&1&1&1&1&1&1&0&1&0&1&1&1&1&0&0&1&0&1  \\
\end{array}
\right]
\end{equation}

\subsection{Noise Model} 
\label{2.2}
The errors that may occur in a single qubit are bit-flip errors (X), phase-flip errors (Z), and amplitude phase errors (Y). In our study, we define the probability distributions of these noises by setting the probability p for the chance of each of these errors occurring in each qubit and $\eta$ as the correlation of the bit-flip errors and phase-flip errors. We also defined the generation probabilities of X-, Y-, and Z-errors, $p_x$, $p_y$, and $p_z$, by using parameters $p$ and $\eta$ as follows: 
\begin{equation}
    p_x = p_z =  \frac{p}{\eta + 2}
\end{equation}
\begin{equation}
    p_y = \frac{\eta p}{\eta + 2}
\end{equation}

Here, $p=p_x+p_y+p_z$, and no error happens at probability $1-p$. Along with the error distributions defined above, we adopted the noise model with $\eta=$0.5, 1, and 3 in the experiments. When $\eta=1$, it follows that $p_x=p_y=p_z$, and the noise becomes the discrete uniform distribution. 
\subsection{Decoder architecture}\label{subsec2}
The present study performed the decoding of the Golay code with the Transformer \cite{vaswani2017} decoder. This decoder treats the decoding procedure as the regression problem depicted below.
\begin{equation}
    \{0,1\}^{22} \to [0,1]^{46}
\end{equation}

The decoder contains only the encoder blocks that output the value between [0,1] for the probability of bit-flip errors or phase-flip errors in 23 qubits from the syndrome measurement made by 22 stabilizer generators. Each output of the Transformer decoder is mapped to 0 or 1 with the boundary set at 0.5. By applying the obtained operators to the error operator giving the correct answer, we can calculate the operator acting on the corrected qubit. The hyperparameters in the decoder are shown in Table~\ref{tab:Training setting}. In each cases, $10^6$ training data and $10^5$ test data were used. 
\begin{table}[h]
\centering
\caption{Training setting}\label{tab:Training setting}
\begin{tabular}{@{}ll@{}}
\hline
Parameters & Values \\
\hline
batch size & $1000$ \\
epoch & $30$ \\
learning rate & $0.0001$ \\
embedding dimension & $128$ \\
number of heads & $8$ \\
number of encoder layers & $4$ \\
loss function & $BCE$ \\
optimizer & RAdam \\
\hline
\end{tabular}
\end{table}

We evaluated the ability of the decoder by measuring the logical qubits. Here, p was varied from 0.1\% to 5\% with a 0.1\% interval, and 10000 decoding experiments were conducted for each p.

\section{Results and discussion} 
\label{3}
\subsection{Effect of the weight of the generator polynomial} 
\label{3.1}
We examined how the decoding accuracy varies under the same noise model for Golay codes 
defined by generator polynomials with weights of 8, 12, and 16. 

\begin{figure}[h]
\centering
\begin{subfigure}{0.32\textwidth}
    \centering
    \includegraphics[width=\linewidth]{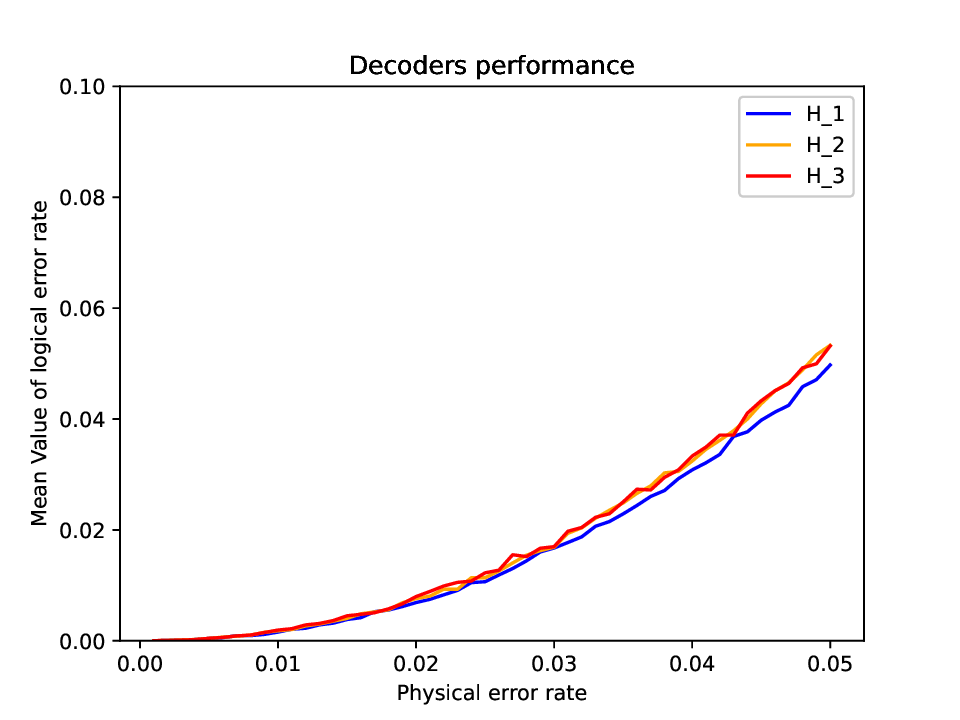}
    \caption{$\eta = 0.25$}
    \label{fig:sub1}
\end{subfigure}
\begin{subfigure}{0.32\textwidth}
    \centering
    \includegraphics[width=\linewidth]{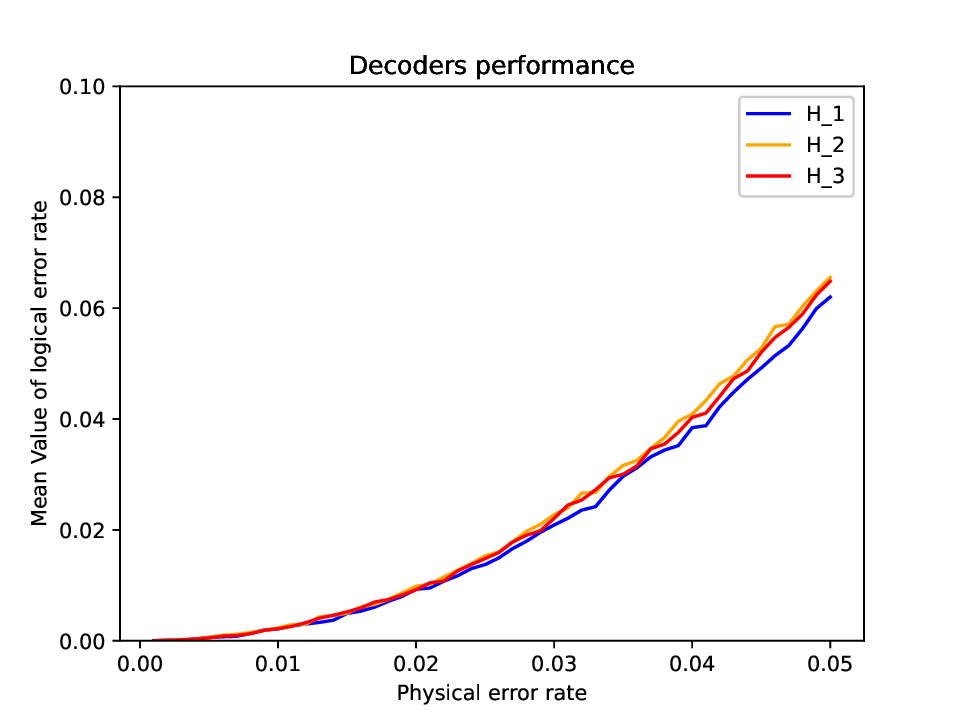}
    \caption{$\eta = 1$}
    \label{fig:sub2}
\end{subfigure}
\begin{subfigure}{0.32\textwidth}
    \centering
    \includegraphics[width=\linewidth]{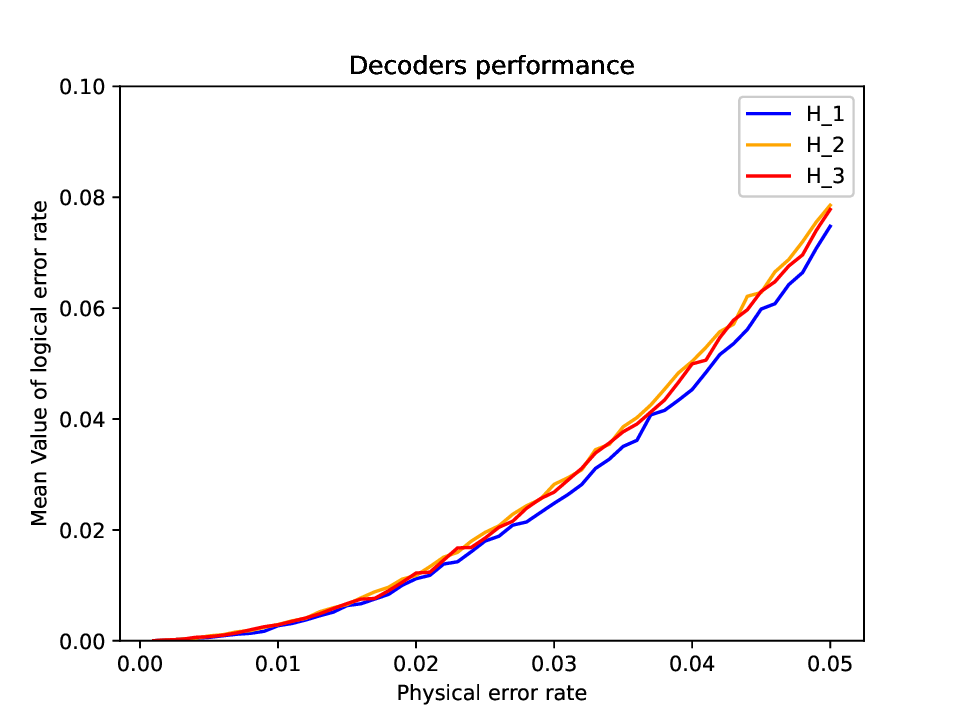}
    \caption{$\eta = 3$}
    \label{fig:sub3}
\end{subfigure}
\caption{Decoder performance of Golay code defined by three generator polynomials for each value of the correlation parameter $\eta$.}
\label{fig:eta}
\end{figure}

For each noise model, Fig.\ref{fig:eta} shows the logical error rates of the Transformer decoders trained to learn the correspondence between errors and syndrome measurement outcomes for the three Golay codes defined by different generator polynomials. 

These results for the same noise model indicate that no significant difference in decoder accuracy depending on the weight of the generator polynomial was observed. The weight of a generator polynomial corresponds to the number of ones in each row of the parity-check matrix, which in turn corresponds to the number of non-identity Pauli operators included in each stabilizer generator. Low-density parity checks are desirable in order to shorten the circuits designed for measurement and thereby reduce decoherence accumulation and measurement overhead. The present results indicate that adopting generator polynomials with smaller weights does not degrade decoding accuracy.

\subsection{Effect of the noise model} 
\label{3.2}
\begin{figure}[h]
\centering
\begin{subfigure}{0.32\textwidth}
    \centering
    \includegraphics[width=\linewidth]{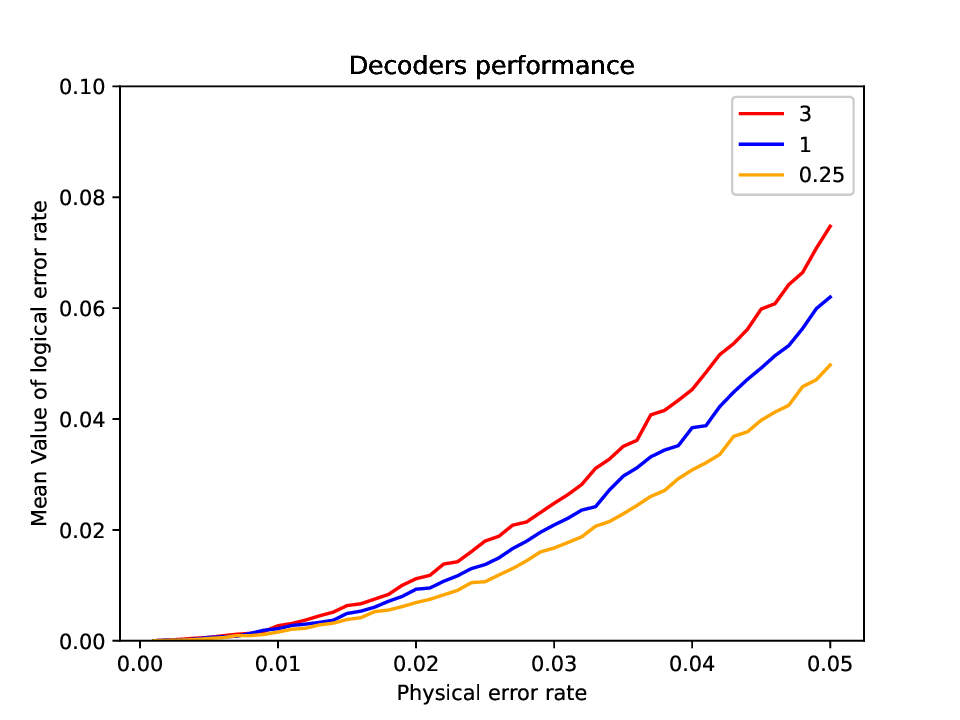}
    \caption{weight = 8}
    \label{fig:sub1}
\end{subfigure}
\begin{subfigure}{0.32\textwidth}
    \centering
    \includegraphics[width=\linewidth]{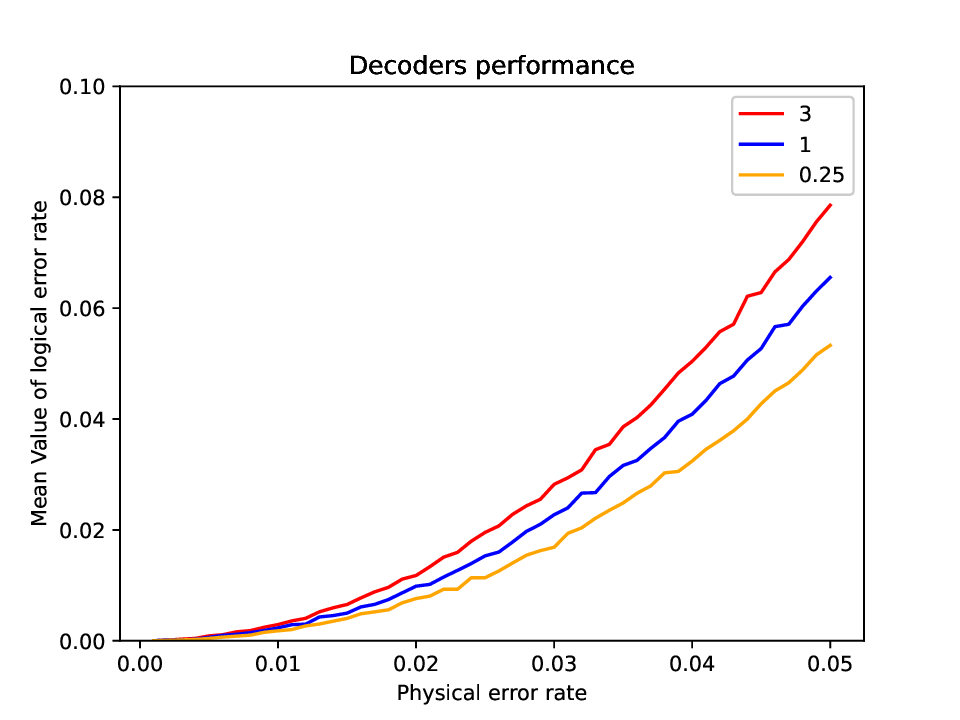}
    \caption{weight = 12}
    \label{fig:sub2}
\end{subfigure}
\begin{subfigure}{0.32\textwidth}
    \centering
    \includegraphics[width=\linewidth]{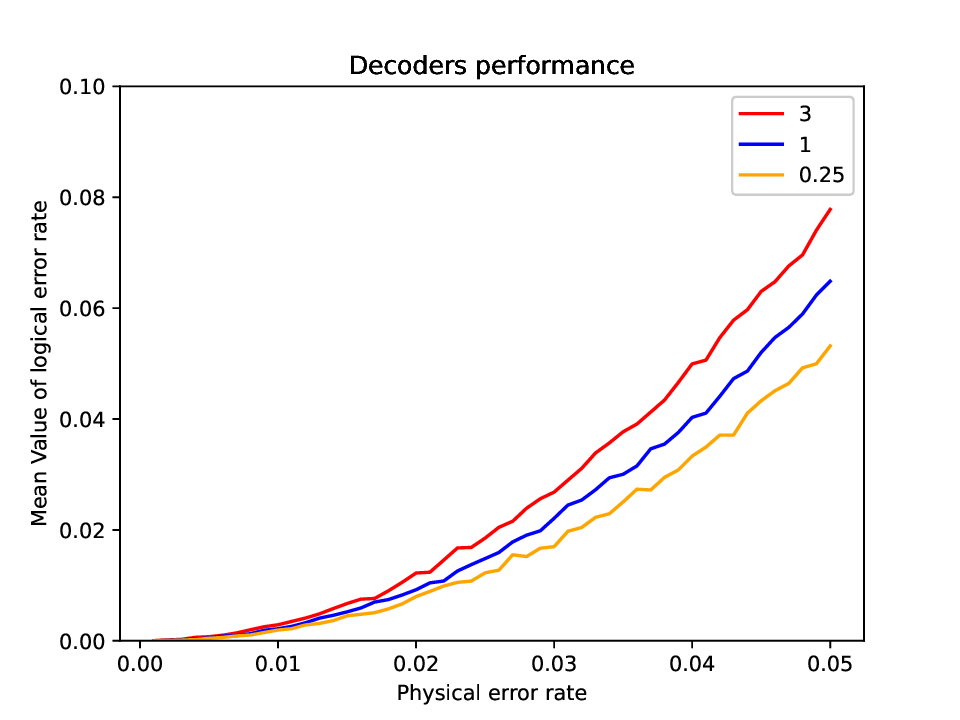}
    \caption{weight = 16}
    \label{fig:sub3}
\end{subfigure}
\caption{Decoder performance of Golay code under three different noise models for each generator polynomial.}
\label{fig:weight}
\end{figure}

For each noise model, we investigated how the decoding accuracy changes for Golay code defined by the same generator polynomial when the parameter $\eta$ was set to 0.25, 1, or 3. For each generator polynomial, Fig.\ref{fig:weight} shows the logical error rates of the Transformer decoders trained to learn the correspondence between errors and syndrome measurement outcomes sampled from the three different noise models. 

The figure shows that, as the parameter $\eta$, which represents the correlation between bit-flip and phase-flip errors, increases, Y-errors become more likely to occur, and when the information source is considered to be $\{I,X,Y,Z\}$, the entropy decreases. However, smaller values of $\eta$ lead to higher decoding accuracy. When the correlation is weaker, the problem of predicting errors from the syndrome measurement outcomes becomes more separable. Since the symbol for Y-errors was not explicitly used in the input or output of the Transformer during training, it is possible that the model was better suited to less correlated noise conditions.

\subsection{Comparison between Golay code and toric code under identical conditions} 
\label{3.3}
\begin{figure}[h]
\centering
\includegraphics[width=0.75\textwidth]{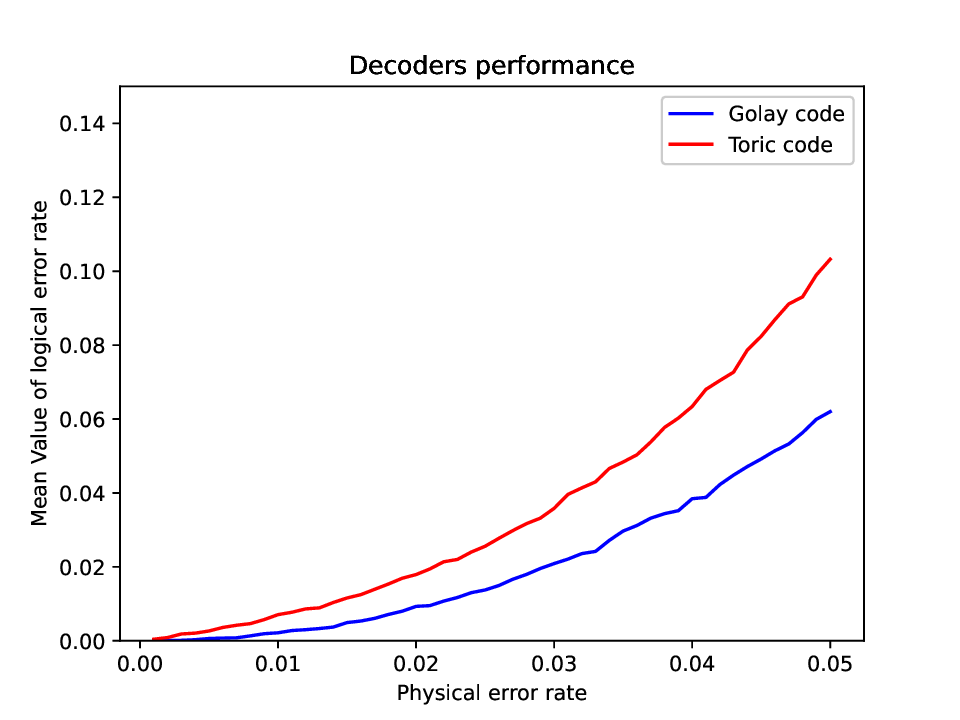}
\caption{Decoder performance of Golay code and toric code (code distance 5) under the noise model with $\eta = 1$.}\label{fig:vstoric}
\end{figure}

We compared the decoding performance of Transformer decoders with identical architectures for Golay code defined by the generator polynomial of weight 8 and toric code with a code distance of 5 under the noise model with $\eta = 1$. Fig.\ref{fig:vstoric} shows the dependence of the logical error rate on the physical error rate for each code. 

The results show that the Golay code consistently achieves a lower logical error rate than that of the toric code. The Golay code had a code distance of 7, which theoretically allows for reliable correction of up to three arbitrary errors, whereas the toric code examined in current study had a code distance of 5, allowing for reliable correction of up to two errors. Considering these differences, the results indicate that the Transformer decoder successfully learned the structural advantage of Golay code in terms of error tolerance. Moreover, since Golay code encodes one logical qubit by using 23 physical qubits, while the toric code encodes two logical qubits using 50 physical qubits, the number of physical qubits required per logical qubit is smaller for Golay code. Taken together, these findings demonstrate that, in quantum error correction using Transformer-based decoding, Golay code outperforms toric code in both error tolerance and encoding efficiency.

\section{Conclusions} 
\label{4}
In current study, we investigated the potential effectiveness of a machine learning–based decoder for [[23,1,7]] quantum Golay code by employing a Transformer decoder composed solely of encoder blocks. The Transformer decoder was trained to learn the correspondence between errors sampled from three noise models with different correlations between bit-flip and phase-flip errors and the resulting syndrome measurement outcomes for three generator polynomials of different weights. The results showed that the weight of the generator polynomial defining the code did not affect the decoder’s accuracy, and that treating the X- and Z-stabilizer syndrome measurements as independent inputs to the model, without explicitly including Y-errors in the output symbols, was an effective strategy when the error correlations were weak. 

Furthermore, under a noise model in which all errors occur with equal probability and with a physical error rate of 5\%, the Transformer decoder for the Golay code achieved a logical error rate of approximately 6\%, representing about a 40\% improvement over that of toric code with code distance 5 under the same conditions. This demonstrates that Golay code surpasses toric code in both error tolerance and qubit efficiency for encoding.

Although the present study employed a simple architecture consisting of only Transformer encoder blocks for training each code, higher accuracy could be expected by designing architectures that better capture the algebraic structure of the Golay code or adapt to specific noise models. Overall, this work highlights the promise of using the Golay code for machine learning–based quantum error correction and suggests significant potential for further research.

\bibliographystyle{unsrt}  
\bibliography{reference}    

\end{document}